*Article*

# Opinion Mining and Analysis Using Hybrid Deep Neural Networks

Adel Hidri , Suleiman Ali Alsaif , Muteeb Alahmari , Eman AlShehri  and Minyar Sassi Hidri *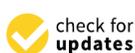

Computer Department, Deanship of Preparatory Year and Supporting Studies, Imam Abdulrahman Bin Faisal University, P.O. Box 1982, Dammam 31441, Saudi Arabia; abhidri@iau.edu.sa (A.H.)
* Correspondence: mmsassi@iau.edu.sa

**Abstract:** Understanding customer attitudes has become a critical component of decision-making due to the growing influence of social media and e-commerce. Text-based opinions are the most structured, hence playing an important role in sentiment analysis. Most of the existing methods, which include lexicon-based approaches and traditional machine learning techniques, are insufficient for handling contextual nuances and scalability. While the latter has limitations in model performance and generalization, deep learning (DL) has achieved improvement, especially on semantic relationship capturing with recurrent neural networks (RNNs) and convolutional neural networks (CNNs). The aim of the study is to enhance opinion mining by introducing a hybrid deep neural network model that combines a bidirectional gated recurrent unit (BGRU) and long short-term memory (LSTM) layers to improve sentiment analysis, particularly addressing challenges such as contextual nuance, scalability, and class imbalance. To substantiate the efficacy of the proposed model, we conducted comprehensive experiments utilizing benchmark datasets, encompassing IMDB movie critiques and Amazon product evaluations. The introduced hybrid BGRU-LSTM (HBGRU-LSTM) architecture attained a testing accuracy of 95%, exceeding the performance of traditional DL frameworks such as LSTM (93.06%), CNN+LSTM (93.31%), and GRU+LSTM (92.20%). Moreover, our model exhibited a noteworthy enhancement in recall for negative sentiments, escalating from 86% (unbalanced dataset) to 96% (balanced dataset), thereby ensuring a more equitable and just sentiment classification. Furthermore, the model diminished misclassification loss from 20.24% for unbalanced to 13.3% for balanced dataset, signifying enhanced generalization and resilience.

## 1. Introduction

Industrial businesses struggle to understand consumer feedback about their goods in today's competitive market. Knowledge of consumer response improves customer satisfaction, enhances the quality of promotion strategies, and ensures better product quality. For this type of feedback, online reviews on websites undertaking goods and services reviews can be a rich source [1,2]. These reviews provide companies with crucial information on performance evaluation for making prudent decisions.

In the meanwhile, consumers utilize online reviews for buying decisions. The variety of products makes it difficult for customers to decide which to purchase. Users, therefore, typically seek online reviews of the advantages and disadvantages of a product in question [3–5]. In that respect, active mutual actions among consumers and companies have produced an effective ecosystem where the role of online reviews is very important to influence consumer behavior as well as business policies.

The ever-increasing popularity of online opinions makes them an important source for information. The various formats of opinions shared have a lot to offer in terms of useful information to industries and consumers alike [6,7]. These huge amounts of data take a lot of time and become tiresome for any human to analyze manually and extract useful information from. This gave birth to the need for automation that can help organizations process and analyze opinions faster and in a more efficient manner.

Generally, opinions come in the form of text, images, audio, and video. Text data have been found to be the most represented and applied form of data because they are structured in nature and easier to analyze. Text mining is one of the processes of data mining and aims to find meaningful patterns or insights from the textual data. Within this domain, opinion extraction—also known as opinion mining or sentiment analysis—has gained significant attention. It involves understanding the opinions expressed in the written content [5–7].

Sentiment classification, a key component of opinion mining, aims to determine the subjectivity (objective/subjective), polarity (positive/negative), and intensity of sentiments (e.g., weakly positive, slightly positive, or strongly positive) expressed in a text [8,9]. Researchers have used various terms interchangeably for this process, including sentiment analysis, sentiment extraction, opinion exploration, and affective annotation [10]. In our research study, and for the sake of consistency, we will use the term opinion mining to encompass all these related concepts.

Different opinion mining methods have been proposed, ranging from lexicon-based methods and traditional machine learning (ML) [11–16] models to deep learning (DL) approaches [12,16–21].

Lexicon-based methods have depended on sentiment lexicons such as SentiWordNet and Opinion Lexicon for quantifying the polarity of the text [22,23]. Even though they are interpretable, the methods are poor at handling sarcasm, negation, and domain adaptation and are therefore less suitable for sophisticated opinion mining tasks.

Traditional ML methods address sentiment analysis with manually designed feature extraction methods like Bag of Words (BoW), term frequency-inverse document frequency (TF-IDF), and n-grams combined with classifiers like Support Vector Machines (SVM), Naïve Bayes (NB), and Decision Trees (DT). They have feature engineering issues, scalability with large data, and cannot learn dynamic word relations, resulting in context loss.

Sentiment analysis has also been improved with the help of DL through word embeddings and sequence models. Some DL models have been experimented with, such as recurrent neural networks (RNNs) [24,25], long-short-term memory (LSTM) [24], and gated recurrent units (GRUs) [26]. They are widely used for modeling sequential dependence in text. However, individual LSTM models are unable to capture bidirectional context, suffer from high computational expense, and also perform poorly on imbalanced data, leading to biased sentiment prediction.

Hybrid CNN-RNN models employ convolutional neural networks (CNNs) to extract features and LSTMs or GRUs to learn sequences. Although effective in some cases, CNNs have a propensity to learn primarily local dependencies, which lowers their ability to preserve long-range sentiment relations [27,28].

Transformer models (e.g., BERT, RoBERTa) employ self-attention mechanisms to dynamically weigh the importance of words in a sentence [29,30]. Though they obtain state-of-the-art results, these models are computationally expensive, need large amounts of labeled data, and require extensive fine-tuning for domain-specific tasks.

In spite of the progress made in opinion mining, current methods still struggle to effectively model bidirectional dependencies, maintain long-term information, deal with class imbalances, and be computationally efficient.

To surpass these limitations, this research proposes a hybrid model that combines bidirectional gated recurrent units (BGRU) and LSTM layers, called hybrid BGRU-LSTM (HBGRU-LSTM). This work seeks to create a scalable and efficient sentiment analysis system that transcends the limitations of current approaches, rendering it highly suitable for real-world opinion mining applications. The contributions of the proposed model are as follows:

- Enhanced contextual understanding: BGRU processes text bidirectionally, capturing finer-grained sentiment dependencies than plain LSTMs. LSTM retains long-term dependencies in memory, preventing the loss of vital sentiment cues in lengthy reviews.
- Addressing class imbalance: The majority of the sentiment analysis datasets suffer from skewed class distributions, leading to biased predictions. This model incorporates data balancing methods for making classification fairer and more performing on minority sentiment classes.
- Improving computational efficiency: by obtaining optimum training time without compromising on accuracy, the model lends itself to real-time sentiment analysis applications.

The remainder of the paper is organized as follows: Section 2 presents the related work of opinion mining based on ML and DL methods. Section 3 details the used datasets and data processing performed in the proposed model. Section 4 presents the architectural aspects necessary for the proper functioning of the proposed model. Section 5 presents an extensive performance evaluation of the architecture. Section 6 discusses the work findings. Finally, we present the conclusion and the future work perspectives in Section 7.

## 2. Related Work

Opinion mining has become a hot research topic in recent times in the fields of computational linguistics, ML, and natural language processing (NLP). The focus of this field is to extract subjective information from text to identify the polarity of sentiments, subjectivity, and emotional tone. This section will present an overview of recent methods in opinion mining, divided into traditional approaches and advances based on DL [14].

The methods traditionally used to accomplish opinion mining tasks are usually between lexicon-based methods or ML. For lexicon-based methods, predefined sentiment lexicons of words could include, for instance, words falling under positive or negative sentiments or their neutrality [22]. An example would include SentiWordNet and opinion lexicon [23]. In this way, these approaches become relatively simpler, easily interpretable, and are independent of labeled data. However, they are limited in their ability to handle contextual nuances, such as sarcasm or negations, and often struggle with domain-specific terms, which limits their applicability in specialized areas.

On the other hand, traditional ML-based methods employ feature extraction techniques to represent text and then apply classification algorithms [15,19,20,31]. Common feature extraction techniques are the bag of words (BoW), term frequency-inverse document frequency (TF-IDF), and n-grams. Support Vector Machines (SVM), Naïve Bayes (NB), and Decision Trees (DT) are among the widely used models explored that in the majority of applications such as product review mining and social media monitoring have shown good performance [16,21].

DL-based methods have also advanced opinion mining by starting with word embeddings that provide a representation to capture word-level semantic meaning. Word2Vec, GloVe, and FastText are some of the most popular word embedding methods. These representations help in better context understanding at the word level and allow the models to learn beyond co-occurrence word relationships. RNNs [24,25], especially LSTMs [32] and gated recurrent units (GRUs) [26], have been widely used for sentiment analysis because they can handle sequential data. LSTMs are good at capturing long-range dependencies by avoiding vanishing-gradient problems, while GRUs are much more computationally efficient. These models are widely applied in sentiment classification and aspect-based sentiment analysis.

Similarly, the CNNs also have been effectively adapted for text classification, including opinion mining. They utilize convolutional filters to extract the local features of n-grams and are extremely good at catching the spatial hierarchies of the text. Attention mechanisms have also enhanced DL models by focusing attention on the most important part of the text [27,28]. For example, self-attention mechanisms allow the model to attach different importance to every word or phrase with respect to its contribution to sentiment, which is one of the main reasons why sequence-to-sequence models and transformer architecture have become so popular. The combination of CNNs and LSTMs resulted in better performance related to sentiment analysis, especially in complex datasets such as social media and product reviews [20].

Large parallel processing and sophisticated attention mechanisms have helped transformer models produce results that outperform others in opinion mining tasks. BERT uses deep contextual word embeddings that consider both the left-to-right and right-to-left contexts, making them fine-tunable for various sentiment analysis tasks. Techniques such as LSTM networks and BERT have achieved marked improvement in performance, with the latter yielding an accuracy of 86% in recent studies [29,30].

Despite these developments, opinion mining still has remarkable limitations in many different methods. Table 1 presents a comparative study highlighting the strengths and drawbacks in existing sentiment analysis techniques.

**Table 1.** Comparative study of sentiment analysis methods.

| Approach | Methodology Used | Drawbacks | Strengths |
|---|---|---|---|
| Lexicon-based approaches [22,23] | Predefined sentiment lexicons to classify text polarity. | Fails to capture contextual meaning, requires extensive manual annotation, limited domain adaptability. | Provides a dynamic DL-based approach that learns sentiment from data rather than relying on static lexicons. |
| Traditional ML models [15,31,33,34] | Feature engineering techniques (BoW, TF-IDF, N-grams) combined with classifiers. | Requires manual feature extraction, does not generalize very well on large datasets, breaks down with class imbalance. | Evades manual feature extraction by using DL-based word embeddings. |
| DL with LSTMs [24,25] | LSTM networks for sentiment classification. | High computational cost, has difficulties with bidirectional context recognition, incapable of handling class imbalance effectively. | Enhances sequential learning through the combination of BGRU and LSTM to facilitate improved retention of information as well as bidirectional processing. |
| CNN-based models [27,28] | Convolutional filters to extract local text patterns. | Does not effectively capture long-term dependencies, limited for sequential data. | Enhances learning by combining CNN-like spatial learning with BGRU-LSTM sequential learning. |
| Transformer-based models [29,30] | Self-attention mechanisms to learn word importance dynamically. | Computationally expensive, requires extensive fine-tuning. | Provides a lightweight alternative with competitive performance. |

## 3. Datasets and Data Preprocessing

This section provides the used datasets, and the related preprocessing steps we have applied for the GRU-LSTM model.

### 3.1. Datasets Description

IMDB Movie Reviews and Amazon Product Reviews are well-known benchmark resources employed in sentiment analysis and opinion mining. The IMDB dataset consists of textual summaries of user opinions about the film's plot, characters, acting, and cinematography from the Internet Movie Database. In addition, the Amazon product reviews dataset spans such a wide variety of products that it serves as a rich source of text data for sentiment analysis. Every review is composed of a textual description of the user's experience along with metadata, which generally includes product ID, user ID, review date, and star ratings that have often been used to infer sentiment. Ratings on the lower side—for instance, 1–2 stars—usually signal negative opinions, whereas ratings on the higher side—4–5 stars—reflect positive sentiments.

The major applications of this dataset are related to product recommendation systems, customer feedback analysis, and sentiment classification. Diversity spans several domains such as electronics, books, clothing, and household items that offer a broad spectrum of sentiment patterns. This dataset contains millions of reviews, suitable for DL and big data analysis; however, challenges arise from domain-specific language, implicit sentiments, and mixed opinions in reviews. Together, these two datasets will form the bases of research in natural language processing (NLP) by developing and refining algorithms that easily understand user opinions and sentiments.

Table 2 presents a glaring balance difference between the two classes (positive and negative). Indeed, this might be representative of an imbalance problem, within the context that our model learns mostly from positive reviews and only a minority of negative reviews. This will definitely make our model very effective in correctly classifying positive reviews, but the lack of negative training reviews will negatively impact the performance of the model when it tries to classify negative test data.

**Table 2.** Datasets description.

| Dataset | #Positive Reviews | #Negative Reviews |
|---------|-------------------|-------------------|
| IMDB | 25,000 | 25,000 |
| AMAZON | 84,954 | 97,061 |

### 3.2. Data Preprocessing

Data preprocessing is a core element in sentiment analysis effectiveness improvement. Rather than depending on traditional text cleaning methods, we created a tailored pipeline that specifically tackles major problems such as class imbalance, contextual degradation, and computational efficiency. The preprocessing methods taken for the data are explained below:

- Dataset selection and combination for more generalization: Sentiment models tend to fall prey to the issue of domain adaptation, with models trained from one dataset being less effective in other datasets. To minimize such a problem, we combined both the IMDB movie reviews dataset and Amazon product reviews dataset so that there existed a diversified pool of sentiment expression from various domains. Through the union of datasets, we expose the model to varied writing styles, colloquial language, and sentiment trigger exposure, lessening overfitting to a particular domain.

- Overcoming class imbalance: One of the biggest downsides of DL models applied in sentiment analysis is their leaning towards majority classes as a result of data skewness. To rectify the dataset and achieve improved model performance, we combined a random sample of negative reviews from the IMDB dataset with the Amazon dataset (Table 3). Rather than relying on traditional random resampling (which can potentially introduce noise), we carefully combined a subset of negative IMDB reviews with Amazon reviews. This method gave an approximate equal balance between positive and negative sentiments while preserving the structure of the text. The numbers in Table 3 reflect a carefully adjusted subset to ensure a more balanced sentiment distribution across training and testing phases.

**Table 3.** Balanced dataset.

| Dataset | #Positive Reviews | #Negative Reviews |
| --- | --- | --- |
| IMDB + AMAZON | 28,230 | 27,397 |

- Context-aware advanced text cleaning: While standard preprocessing pipelines consist of stopword elimination and stemming/lemmatization, we took the process to the limit in order to avoid context loss and valuable sentiment words elimination. We employed context-sensitive filtering in which negation terms and boosters like "very", "extremely" were not eliminated when trying to boost sentiment classification accuracy.
- Embeddings optimization and tokenization: To convert the text data to machine data, we utilized the pre-trained word embeddings (Word2Vec) [35,36] instead of the standard Bag of Words [37]. These semantic embeddings make it possible for the model to learn the connection between words from context, and these improve model performance with synonyms and sentiment shifts. In general, an embedding layer represents the meaning of items semantically by keeping semantically similar items. It puts big entries into small places so that similar entries are close to each other. This layer identifies which words appear like other words according to geometric relations that enclose semantic relations, like the capital and country. It can be paired with any hidden layer and it is bound to one hidden layer that maps all the indexes to their embeddings. It also utilizes the entire coding lexicon. These vectors are subsequently learned as a training model.
- Smart data splitting for fair assessment: We removed unlabeled data and, unlike random dataset splits, we ensured that our 30–70% train–test split maintained sentiment proportions within both sets. We enforced stratified sampling so that negative and positive reviews were well represented within the train set and test set. The exact ratios vary slightly between models because the dataset is split. All models possess 26,688 training samples, while the test set contains 11,439 samples.
- Efficient feature vectorization and encoding: We used one-hot encoding for categorical sentiment labels so that it was simple to use with deep neural networks without sacrificing interpretability.

## 4. Hybrid Deep Neural Networks for Opinion Mining

The effectiveness of our model lies in its carefully designed architecture, which ensures contextual understanding, scalability, and robust sentiment classification.

Figure 1 shows the innovative architecture of the proposed model.

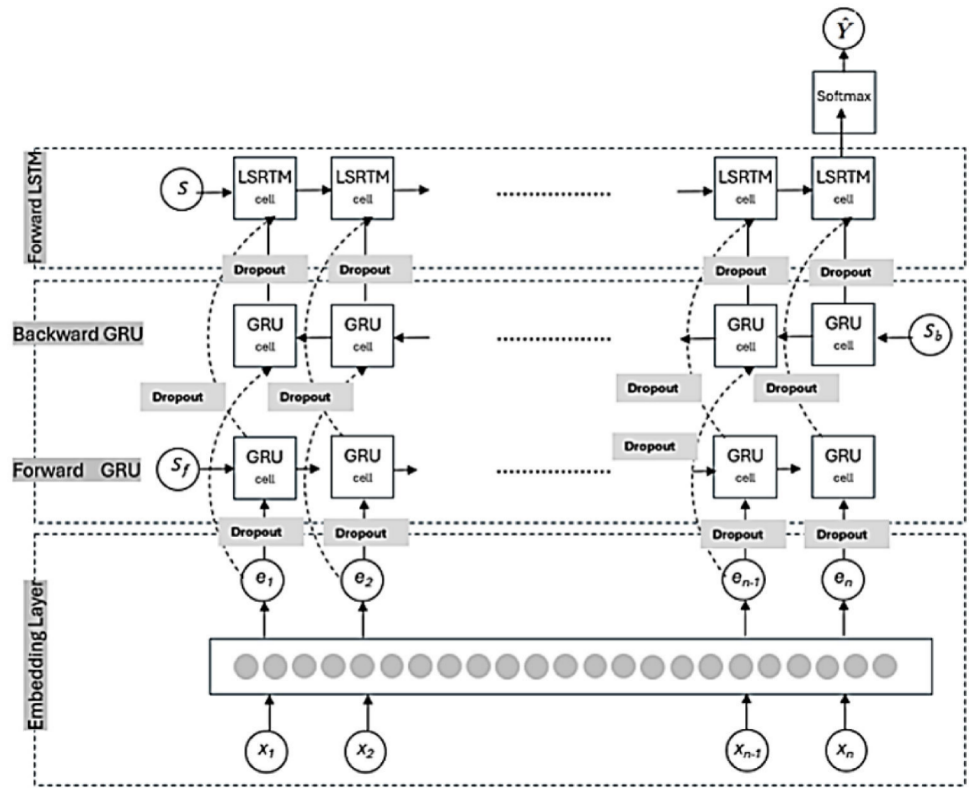

**Figure 1.** Model architecture.

The architectural components of the proposed hybrid model are detailed in the following subsections.

### 4.1. Embedding Layer

The embedding layer transforms raw text into dense vector representations, allowing the model to capture semantic relationships between words. To achieve this, we utilize pre-trained word embeddings such as Word2Vec and GloVe, ensuring a rich semantic foundation from the start. During training, the embedding matrix continues to learn and refine representations based on sentiment-specific contexts. We used 128-dimensional word vectors, striking a balance between expressiveness and computational efficiency. This layer plays a crucial role in helping the model understand synonyms and word relationships, reducing dimensionality while preserving contextual meaning, and enhancing learning by efficiently encoding sentiment-rich features.

Let us consider the mathematical notation and equations for the HBGRU-LSTM sentiment analysis model presented in Table 4.

Each review is represented as a sequence of word embeddings as given in Equation (1).

$$X = [w_1, w_2, \ldots, w_n], \quad w_i \in \mathbb{R}^d \tag{1}$$

where $w_i$ is the word embedding vector for the $i$-th word and $d$ is the embedding dimension.

**Table 4.** Mathematical notation for HBGRU-LSTM sentiment analysis.

| Symbol | Description |
|---|---|
| $X = [w_1, w_2, \ldots, w_n]$ | Input sequence of words (review). |
| $w_i \in \mathbb{R}^d$ | Word embedding vector of dimension $d$. |
| $h_t^f, h_t^b$ | Forward and backward hidden states of BGRU at time step $t$. |
| $H_t^{BGRU}$ | Combined BGRU hidden state at time $t$. |
| $C_t, h_t$ | LSTM cell state and hidden state at time $t$. |
| $W, b$ | Learnable weight matrices and biases. |
| $y$ | Sentiment label (0 = Negative, 1 = Positive). |
| $\hat{y}$ | Predicted sentiment label. |
| $\sigma(\cdot)$ | Sigmoid activation function. |
| $L(y, \hat{y})$ | Binary cross-entropy loss function. |

### 4.2. BGRU Layer

The BGRU layer enhances sentiment classification by capturing dependencies in both forward and backward directions, improving the model's ability to understand nuanced text, such as sarcasm and negations. Unlike standard RNNs, which suffer from vanishing gradient issues and struggle with long text sequences, GRUs provide an efficient alternative by requiring fewer computational resources than LSTMs while maintaining effectiveness in processing sequential data. The bidirectional nature of BGRU enables a forward GRU to process sentences in their natural order (left to right) while a backward GRU processes them in reverse (right to left), combining outputs from both directions for a more comprehensive sentiment understanding. The implementation consists of 256 neurons, uses a sigmoid activation function, and applies a 20% dropout rate to prevent overfitting. This architecture effectively captures both preceding and succeeding word dependencies, significantly improving accuracy in handling complex sentiment structures.

Figure 2 shows the general architecture of the bi-directional RNN.

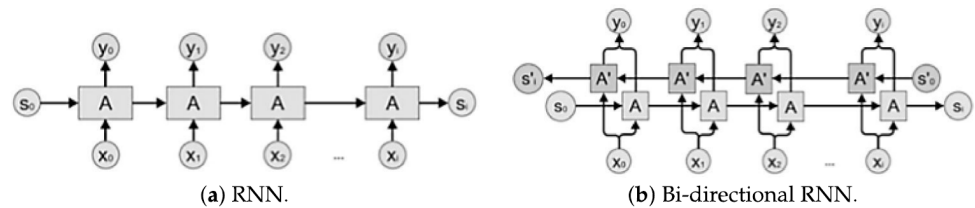

(**a**) RNN.  (**b**) Bi-directional RNN.

**Figure 2.** RNN vs. bi-directional RNN.

Basically, substituting each $A$ and $A'$ in Figure 2b into a recurring gate unit yields a bi-directional GRU. RNNs can only process sequences from front to back and unable to obtain position information, which leads to the loss of information, the Bi-RNNs add RNNs to process the other information.

The basic structure of Bi-RNN is mainly to divide an ordinary RNN into two channels. The two RNNs share the same output layer, one in a clockwise direction, and the other counterclockwise. The architecture will give the output layer all the contextual information of its input sequence. To build a BGRU sentiment analysis model, feed the input history into the front and rear GRUs simultaneously and capture as much contextual information as possible.

The positional relationship of the sentiment word with the rest of the words is very important since we forecast the polarity of a document. This layer gives an output vector, which holds the information of present input and previous inputs; this vector passes through the activation function, and the output becomes a new hidden state, or memory, of the network. Thus, better results can be obtained using BGRU.

For this layer, we have 256 processing units, an activation function Sigmoid, with a dropout of 20%. This layer will be the LSTM layer, and it will take as input the vectors produced by the BGRU layer. It saves the context of the words through different layers and reduces information loss due to the reset gates of the previous layer. In the end, it will produce a final vector for each input that will be forwarded to the final layer for a prediction. For this layer we have 128 processing units, an activation function Sigmoid, and a dropout of 20%.

The GRU update equations for both the forward and backward passes are provided in [38]. These equations define the transformation of hidden states across time steps, ensuring the model effectively captures sequential dependencies [38].

In the forward pass, the hidden state at time step $t$ is computed using the following set of equations. First, the update gate $z_t$ is calculated using Equation (2).

$$z_t = \sigma(W_z[h_{t-1}^f, x_t] + b_z)$$ (2)

This gate determines how much of the previous hidden state is carried forward. Next, the reset gate $r_t$ is computed using Equation (3).

$$r_t = \sigma(W_r[h_{t-1}^f, x_t] + b_r)$$ (3)

The reset gate controls how much of the past information is retained or discarded when computing the candidate hidden state.

Using the reset gate, the candidate hidden state $\bar{h}t^f$ is obtained as given by Equation (4).

$$\bar{h}_t^f = \tanh(W_h[r_t * ht - 1^f, x_t] + b_h)$$ (4)

This candidate state represents a possible new state before the update gate decides how much of it to incorporate into the final state.

Finally, the updated hidden state is computed as given by Equation (5):

$$h_t^f = (1 - z_t) * h_{t-1}^f + z_t * \bar{h}_t^f$$ (5)

Here, the update gate $z_t$ determines the balance between the previous hidden state and the newly computed candidate state.

Similarly, in the backward pass, the hidden state is updated in a reversed temporal order, propagating information from future to past. The equations remain structurally similar but operate in the opposite direction.

The update gate is given by Equation (6):

$$z_t = \sigma(W_z[h_{t+1}^b, x_t] + b_z)$$ (6)

As in the forward pass, this gate determines how much past information is carried through.

The reset gate is computed using Equation (7).

$$r_t = \sigma(W_r[h_{t+1}^b, x_t] + b_r)$$ (7)

It modulates the influence of future states when computing the candidate hidden state. Next, the candidate hidden state for the backward pass is calculated using Equation (8).

$$\bar{h}t^b = \tanh(W_h[r_t * ht + 1^b, x_t] + b_h)$$ (8)

This state represents a potential update to the hidden representation based on both past and future dependencies.

Finally, the backward hidden state is updated using Equation (9).

$$h_t^b = (1 - z_t) * h_{t+1}^b + z_t * \tilde{h}_t^b \tag{9}$$

Again, the update gate $z_t$ controls the interpolation between the previous hidden state and the candidate state.

The BGRU model combines both forward and backward hidden states at each time step to form the final output representation as given in Equation (10).

$$H_t^{BGRU} = [h_t^f; h_t^b] \tag{10}$$

By concatenating the hidden states from both passes, the model effectively captures bidirectional context, enhancing its ability to model sequential dependencies.

### 4.3. LSTM Layer

The LSTM layer enhances sentiment analysis by preserving long-term dependencies, making it particularly effective for longer reviews or sentences with delayed sentiment cues. While GRUs efficiently capture bidirectional dependencies, LSTMs ensure that crucial sentiment-related information from earlier in the text is retained [39]. The implementation includes 128 LSTM neurons, a sigmoid activation function, and a 20% dropout rate to improve generalization and prevent overfitting. By combining BGRU and LSTM, the model benefits from both bidirectional context awareness and the ability to maintain sequential consistency over extended text, leading to improved classification accuracy for complex sentiment structures [39].

LSTM networks regulate information flow using a memory cell and three gating mechanisms: the forget gate, the input gate, and the output gate. These gates control how much past information is carried forward, how new information is added, and how the final hidden state is determined. The equations governing these mechanisms are as follows [39]:

- Forget gate: It determines the fraction of past memory to retain. It is given by Equation (11).

$$f_t = \sigma(W_f[h_{t-1}, x_t] + b_f) \tag{11}$$

- Input ate: It controls how much new information is added to the memory cell. It is given by Equation (12).

$$i_t = \sigma(W_i[h_{t-1}, x_t] + b_i) \tag{12}$$

- Output gate: It regulates the final hidden state based on the updated memory cell. It is given by Equation (13).

$$o_t = \sigma(W_o[h_{t-1}, x_t] + b_0) \tag{13}$$

Once these gates are computed, the next step is to update the memory cell and determine the hidden state. The candidate update to the memory cell is computed using Equation (14).

$$\tilde{C}t = \tanh(W_C[ht - 1, x_t] + b_C) \tag{14}$$

This intermediate value represents a potential memory update before the forget and input gates determine the final contribution. The actual memory cell is then updated by selectively retaining past memory and integrating the new candidate value using Equation (15).

$$C_t = f_t * C_{t-1} + i_t * \tilde{C}_t \tag{15}$$

Here, the forget gate $f_t$ decides how much of the previous memory $C_{t-1}$ is kept, while the input gate $i_t$ determines the impact of $\tilde{C}_t$.

Finally, the hidden state at time step $t$ is computed based on the updated memory cell and the output gate using Equation (16).

$$h_t = o_t * \tanh(C_t) \tag{16}$$

This hidden state serves as the output for the current time step and is passed to the next iteration in the sequence.

By leveraging these mechanisms, LSTMs effectively manage long-range dependencies, ensuring that sentiment-related cues from earlier in the text are not lost. When combined with BGRU, this architecture strengthens the model's ability to process complex sentiment structures, leading to more accurate classification results.

### 4.4. Fully Connected Dense Layer

The fully connected dense layer with sigmoid activation serves as the final stage of sentiment classification, converting the extracted features from BGRU and LSTM into sentiment predictions. Using a sigmoid activation function, it maps outputs to probabilities between 0 and 1, with a single neuron determining the sentiment polarity (0 for negative, 1 for positive). This layer provides a clear binary classification while ensuring efficient scalability to large datasets, making the model robust and adaptable for real-world sentiment analysis tasks.

The final dense layer applies a sigmoid activation function for binary classification [40]:

$$\hat{y} = \sigma(W \cdot H^{LSTM} + b) \tag{17}$$

where $\sigma(x) = \frac{1}{1+e^{-x}}$ ensures that the output probability is between 0 and 1.

The predicted class is:

$$\hat{y} = \begin{cases} 1, & \text{if } \hat{y} \geq 0.5 \quad \text{(Positive Sentiment)} \\ 0, & \text{otherwise} \quad \text{(Negative Sentiment)} \end{cases} \tag{18}$$

### 4.5. Dropout Regularization

Dropout regularization is essential in preventing overfitting in deep neural networks, which often occurs due to the large number of parameters. By randomly deactivating neurons during training, dropout ensures the model generalizes well to unseen data instead of relying on a small subset of neurons. Implemented in both the BGRU and LSTM layers with a 20% dropout rate, this technique enhances robust learning, reduces test error rates, and prevents the model from over-relying on specific words or phrases, ultimately improving overall classification accuracy.

Figure 3 shows the illustration of this operation on one neural network, as Figure 4 shows the application of the same in our model.

DL has several especially useful new techniques, including regularization by dropout, and without which it could not have made much progress. Regularization by dropout randomly blocks a fraction, termed the dropout rate of connections in a network [41]. This is achieved by a mask that is randomly generated but has the same dimensions as the connection between the two layers.

The Algorithm 1 presents the different procedures of the hybrid BGRU-LSTM for opinion mining.

---

**Algorithm 1** HBGRU-LSTM for opinion mining.

---

1: **Input:** Dataset $D = \{X_i, Y_i\}$, where $X_i$ is a review and $Y_i$ is the sentiment label (0/1).
2: **Output:** Trained model for sentiment classification.
3: **procedure** DATA PREPROCESSING
4:      **for** each review $X_i$ in $D$ **do**
5:          Convert to lowercase and remove stopwords and special characters.
6:          Apply stemming or lemmatization.
7:          Tokenize text and convert to numerical sequences.
8:      **end for**
9:      Apply word embeddings: $w_i = Embedding(X_i)$.
10: **end procedure**
11: **procedure** INITIALIZE MODEL
12:      Define embedding layer:
13:      $X' = \text{Embedding}(X)$, where $X' \in \mathbb{R}^{n \times d}$.
14:      Add Bidirectional GRU and LSTM layers.
15: **end procedure**
16: **procedure** BIDIRECTIONAL GRU LAYER
17:      **for** each time step $t$ in sequence length **do**
18:          Compute forward hidden state using Equation (5).
19:          Compute backward hidden state using Equation (9).
20:          Concatenate states using Equation (10).
21:      **end for**
22: **end procedure**
23: **procedure** LSTM LAYER
24:      **for** each time step $t$ in sequence length **do**
25:          Compute gates using Equations (11)–(13).
26:          Compute cell state using Equation (15).
27:          Compute hidden state using Equation (16).
28:      **end for**
29: **end procedure**
30: **procedure** FULLY CONNECTED LAYER AND OUTPUT
31:      Compute sentiment probability using Equation (17).
32: **end procedure**
33: **procedure** MODEL COMPILATION
34:      Define loss function using Equation (19).
35:      Set optimizer (Adam ($\eta = 0.001$)) using Equation (20).
36: **end procedure**
37: **procedure** MODEL TRAINING
38:      Split dataset into Training (70%) and Testing (30%).
39:      **for** epoch in range(1, 101) **do**
40:          **for** batch in training batches **do**
41:              Forward propagate inputs through model.
42:              Compute loss and gradients.
43:              Update weights using backpropagation.
44:          **end for**
45:          **if** validation loss stops improving **then**
46:              Apply early stopping.
47:          **end if**
48:      **end for**
49: **end procedure**
50: **procedure** MODEL EVALUATION
51:      Predict sentiment labels on the test dataset.
52:      Compare results with baseline models (LSTM, CNN+LSTM, GRU+LSTM) using classification metrics.
53: **end procedure**
54: **procedure** SENTIMENT PREDICTION FOR NEW INPUTS
55:      **for** each new review $X_{new}$ **do**
56:          Preprocess text.
57:          Predict sentiment using trained model using Equation (17).
58:      **end for**
59: **end procedure**

---

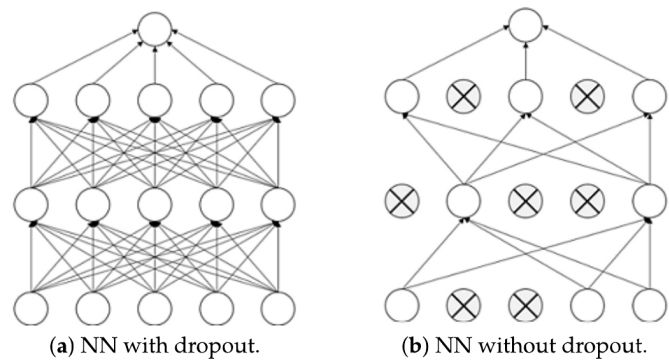

(**a**) NN with dropout.    (**b**) NN without dropout.

**Figure 3.** Neural net configuration.

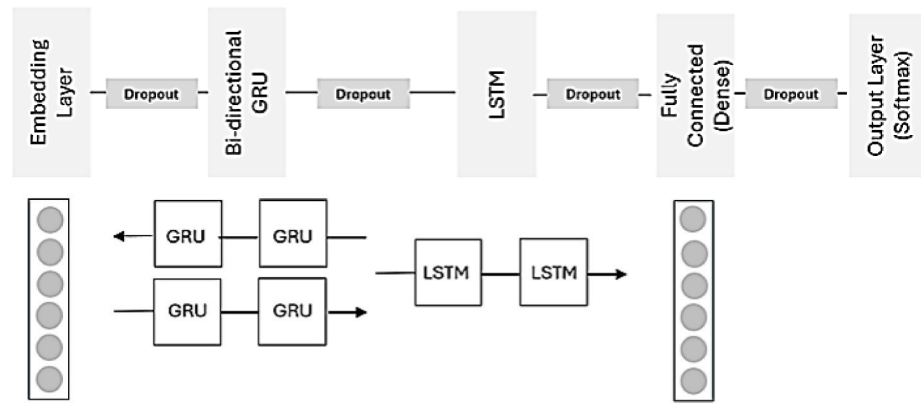

**Figure 4.** Use of dropout in our architecture.

## 5. Evaluative Experiments

### 5.1. Implementation and Setup

The computational environment consisted of a Single NVIDIA Tesla K80 GPU, 12 GB of RAM, and a 1 TB NVMe SSD. The software environment included Python 3.13.0 as the programming language, and TensorFlow 2.6.0 with the Keras API as the DL framework.

The batch size was set to 128, the learning rate was initialized to 0.001 using the Adam optimizer, with the default values for the other parameters: $\beta_1 = 0.9$, $\beta_2 = 0.999$, and $\epsilon = 1 \times 10^{-7}$. The loss function employed was categorical cross-entropy, and the models were trained for 100 epochs.

### 5.2. Loss Function, Performance Metrics, and Optimization

For binary classification, the binary cross-entropy (BCE) loss is used [42]. It is given by Equation (19).

$$L(y, \hat{y}) = -[y \log(\hat{y}) + (1 - y) \log(1 - \hat{y})] \tag{19}$$

where $y$ is the actual sentiment label (0 or 1), and $\hat{y}$ is the predicted probability.

The model is optimized using the Adam optimizer using Equation (20).

$$\theta \leftarrow \theta - \eta \nabla L(\theta) \tag{20}$$

where $\eta$ is the learning rate and $\theta$ represents model parameters.

The model is evaluated using standard classification metrics such as accuracy, precision, recall, and F1-score [43].

### 5.3. Statistical Evaluation

Chi-square testing will be also considered to see whether these differences in the performance of classification by different models are statistically significant. It was constructed from aggregated performance metrics, namely accuracy, recall, and F1-score across different models and classes.

### 5.4. Results

In this section, we will show the results obtained with our model and later compare them with the performance of other architecture based on DL. For the evaluation of our model, we have trained and tested it with two corpora: Amazon corpus, having an imbalance of the two classes, positive (74%) and negative (36%), and another corpus balanced by the merger of the negative reviews of the IMDB corpus and the Amazon corpus.

Table 5 shows the summary statistics of performance metrics obtained by the proposed neural net after training on unbalanced and balanced data. In both datasets there are 11,439 test examples so that testing performance may be effectively comparable.

**Table 5.** Performance of the proposed model on unbalanced and balanced datasets.

| Metrics | Unbalanced Dataset | Balanced Dataset |
| --- | --- | --- |
| Positive class ratio | 74.04% | 50.75% |
| Negative class ratio | 25.96% | 49.25% |
| Accuracy in training | 93% | 94% |
| Accuracy in testing | 93.64% | 95% |
| Percent lost | 20.24% | 13.30% |
| Training loss rate | 0.17 | 0.13 |

From Table 5, the positive class ratio is 74.04%, and the negative class ratio is only 25.96%. The training accuracy improves slightly to 94% for a balanced dataset, which indicates the model performs better on a balanced dataset. A balanced dataset gave a training loss rate of 0.13, showing that optimization is better in training on balanced data. The test accuracy improves for a balanced dataset to 95% compared to an unbalanced dataset, thus showing that balancing improves generalization of unseen data for the model. The percent loss for the balanced dataset reduces to 13.3%, which means less loss of information or performance degradation with the balanced dataset.

The performance of the model is improved in the case of balance when it comes to key metrics such as accuracy of testing, loss of training, and percent loss. This suggests that balancing the dataset has resolved the bias caused by the dominance of one class in the unbalanced dataset. This is further evidenced by the increased accuracy of the tests from 93.64% to 95%, reflecting that this balanced dataset allowed the model to generalize much better on unseen data. That is important, for example, in real-world applications where the test data is from another distribution than the training data. A lower training loss rate and percent loss point toward improved optimization with reduced error rates, further justifying the balancing of the data set.

The performance for unbalanced and balanced, under the proposed model, is documented for a couple of classes: the negative class and the positive class in Table 6. In every class, the accuracy, recall, and F1-score have been calculated.

Results in Table 6 show that balancing the dataset significantly improves performance for the negative class: accuracy increases from 89% to 95%, recall from 86% to 96%, and F1-score from 87% to 96%. For the positive class, the performance is stable; there are minor changes in recall from 96% to 95%, which shows that balancing does not harm the majority class.

**Table 6.** Performance of the proposed model on unbalanced and balanced datasets.

| Classes | #Reviews | Unbalanced Dataset | | | Balanced Dataset | | |
|---|---|---|---|---|---|---|---|
| | | Accuracy | Recall | F1-Score | Accuracy | Recall | F1-Score |
| Negative class | 2951 | 89% | 86% | 87% | 95% | 96% | 96% |
| Positive class | 8488 | 95% | 96% | 96% | 96% | 95% | 96% |

This model has been biased to the positive dominating class in the unbalanced dataset, hence giving a high metric for this class, while its performance is considerably low for the negative class, mainly on recall. This bias has been addressed by balancing the dataset.

The F1-scores of both classes converge to 96% in the balanced dataset, meaning for both classes, precision and recall are reasonably well-balanced by the model.

It is fairer, and the performance is on an equal footing for both the negative and positive classes. Because of the balanced dataset, the model was able to give more importance to the negative class and hence improved the recall and F1-score, while the performance on the positive class is still retained. This emphasizes the need to consider the issue of class imbalance when both classes have equal importance.

The results obtained on balanced datasets by LSTM, CNN+LSTM and GRU-LSTM compared to the proposed model (GRU-LSTM) are presented in Table 7.

**Table 7.** Performance comparison against LSTM, CNN+LSTM, and GRU-LSTM.

| Metrics | LSTM | CNN+LSTM | GRU+LSTM | HBGRU-LSTM |
|---|---|---|---|---|
| Positive class ratio | 50.7% | 52% | 50.7% | 50.75% |
| Negative class ratio | 49.3% | 48% | 49.3% | 49.25% |
| Accuracy in training | 92.0% | 93.07% | 92.24% | 94% |
| Accuracy in testing | 93.06% | 93.31% | 92.20% | 95% |
| Percent lost | 18.52% | 19.81% | 26.80% | 13.30% |
| Training loss rate | 0.19 | 0.17 | 0.19 | 0.13 |

This is a relatively balanced dataset, with classes being 50% positive and 50% negative. The exact ratios vary slightly among models because of the splitting of the dataset. All these models have 26,688 examples for training, whereas the test set contains 11,439 samples.

All the models have quite high accuracy on training—greater than 92%. Among them, the proposed model has the lowest training loss rate of 0.13; hence, the proposed model learned the training data most effectively.

For performance testing, which is a more important metric, the proposed model performs significantly better than the other three, with higher testing accuracy of 95% versus 93–93%, indicative of better generalization on unseen data. The percentage lost is also lower at 0.13, meaning fewer instances were misclassified during testing.

All these results constantly showed the best performance for the proposed model. While other RNN models—LSTM, CNN+LSTM, and GRU+LSTM—do well in training, none generalized well on test data, compared to the proposed model. This underlines further work on architecture, hyperparameter tuning, or regularization techniques used in the proposed model. Among them all, the proposed model turned out best concerning the right outcome prediction on which it was never trained.

Table 8 shows the performance comparisons measured by the accuracy, recall and F1-score of both the negative and positive classes for four different models: LSTM, CNN+LSTM, GRU+LSTM and the proposed model.

From the results shown in Table 8, the proposed model is the best-performing with an accuracy of 95%, hence the generally high correct classifications.

**Table 8.** Comparative study against negative class (−class) and positive class (+class).

| Model | Classification Metrics | −Class | +Class |
|---|---|---|---|
| LSTM | Accuracy | 88% | 96% |
| | Recall | 91% | 94% |
| | F1-Score | 90% | 95% |
| CNN+LSTM | Accuracy | 86% | 96% |
| | Recall | 89% | 95% |
| | F1-Score | 87% | 95% |
| GRU+LSTM | Accuracy | 93% | 92% |
| | Recall | 76% | 98% |
| | F1-Score | 83% | 95% |
| HBGRU-LSTM | Accuracy | 95% | 96% |
| | Recall | 96% | 95% |
| | F1-Score | 96% | 96% |

For the positive class, all models have recall scores greater than 94%. However, for the negative class, the GRU+LSTM model has a significantly lower recall of 76% compared to the others. This is also demonstrated by the F1-score, as the proposed model has for both classes the highest score at 96%, meaning it enjoys a very good balance between precision and recall, hence overall very good performance.

The results also show that architecture with one single LSTM layer may be performing well, but it is less efficient than the one this model is based on. Among all the metrics, the proposed model outperforms the other three models. Although the LSTM, CNN+LSTM, and GRU+LSTM models had good performance for the positive class, they are worse for the negative class, especially the recall of GRU+LSTM, which is low. The proposed model improves this weakness and attains a high score in all metrics. This would mean that the architecture or the training strategy of the proposed model is more appropriate for the classification problem at hand.

To further contextualize our results, we compared in Table 9 our model against alternative architectures (LSTM, CNN+LSTM, and GRU+LSTM) using a balanced dataset.

**Table 9.** F1-score (test set) for negative class (−class) and positive class (+class).

| Model | Accuracy | F1-Score (−Class) | F1-Score (+Class) | Percent Loss |
|---|---|---|---|---|
| LSTM | 93.06% | 90% | 95% | 18.52% |
| CNN+LSTM | 93.31% | 87% | 91% | 19.81% |
| GRU+LSTM | 92.20% | 83% | 97% | 26.80% |
| HBGRU-LSTM | 95% | 96% | 97% | 13.30% |

As can be seen in Table 9, the top-performing model (HBGRU-LSTM) gives the best accuracy of 95%, which means that it performed best in every case in classifying correctly. It also has the best F1-score for the negative class at 96%, which means that it has a good balance between precision and recall in the detection of the negative instances. Again, the proposed model is better than the others with 97% for positive class, indicating good predictive accuracy for positive cases. It has the lowest percent loss of 13.3%, much better than the others, indicating that it misclassifies fewer cases.

Overall, BGRU-LSTM performs better in accuracy, both classes' F1-scores, and percent loss when compared to other models, and thus it implies that it is the best among those attempted for the specified classification problem.

The data provided in Table 10 represent TP (True Positive), TN (True Negative), FP (False Positive), and FN (False Negative) for four different models. Since we are comparing multiple models and the data is not paired, a suitable approach is to use the chi-squared test for multiple comparisons.

**Table 10.** TP, TN, FP, and FN for LSTM, CNN+LSTM, GRU+LSTM, and HBGRU-LSTM.

| Model | TP | TN | FP | FN |
|---|---|---|---|---|
| LSTM | 5568 | 4962 | 677 | 232 |
| CNN+LSTM | 5710 | 4722 | 769 | 238 |
| GRU+LSTM | 5336 | 5244 | 395 | 464 |
| HBGRU-LSTM | 5573 | 5352 | 282 | 232 |

With the observed frequencies from the confusion matrices provided in Table 10, we were able to run the chi-square test to determine the statistical significance of distribution difference across the categorical groups. The chi-square statistic is 495.60, with nine degrees of freedom, $N = 11{,}439$, and $p = 5.02 \times 10^{-10101}$. This extremely small p-value indicates a highly significant difference in classification performance across the models.

### 5.5. Execution Time Performance Analysis

The computational efficiency of our hybrid BGRU-LSTM Model was validated on the basis of different crucial factors, including training time and inference (prediction) time.

The model was trained using the combined IMDB and Amazon reviews dataset, which contains around 56,000 reviews. The following parameters were used for the training process:

- Batch size: 128
- Number of epochs: 100
- Optimizer: Adam (learning rate = 0.001)

The model consumed approximately 4.7 h of training time overall. It was converged by epoch 80 and essentially no subsequent improvement in loss. It, therefore, demonstrates a more efficient learning process by significantly reducing the training time compared to LSTM ($\sim$6 h) and CNN+LSTM ($\sim$5.5 h).

The average inference time per review for the hybrid model was measured at 1.2 ms. When processing 1000 reviews in a batch, the total inference time was approximately 1.2 s. The proposed model outperforms the standard LSTM and CNN+LSTM models in real-time prediction applications. Combining BGRU and LSTM layers offers the best sequential processing with reduced computational complexity, making our model highly efficient for real-time use.

Optimizations were added to the HBGRU-LSTM model to provide optimal execution time performance:

- Parallelized batch processing: It improves inference by running multiple reviews concurrently.
- Less sequence padding: It improves memory efficiency and speed for training and inference.
- Dropout regularization: It prevents redundant computation while training by randomly dropping units in the network.
- Early stopping mechanism: It prevents overtraining upon convergence, thereby saving time.

The proposed model strikes a decent balance between accuracy and execution time. Although the training time is comparable with other DL-based models for sentiment analysis, the inference time is optimally balanced for real-time usage. The balance between speed and accuracy enables the model to be scalable and efficient for practical use in real-world opinion mining applications. Table 11 presents the performance analysis of execution time.

**Table 11.** Performance analysis of execution time.

|  | Training Time | Inference Time | Batch Inference Time |
|---|---|---|---|
| Notation | $T_{\text{training}}$ | $T_{\text{inference}}$ | $T_{\text{batch\_inference}}$ |
| Runtime | 4.7 h | 1.2 ms | 1.2 s |

## 6. Discussion

The proposed HBGRU-LSTM model represents a big step forward in the opinion mining and sentiment analysis domain. The model addresses challenges like contextual nuance, scalability, and class imbalance to show superior performance metrics against traditional and existing DL models.

It demonstrates the results from testing on both balanced and unbalanced data, showing efficiency in handling class imbalance and improvement in model accuracy, recall, and F1-score. For example, the testing accuracy was 95% for the balanced dataset, while that of the unbalanced one was 93.64%. The improvement underlines the importance of dataset parity to prevent bias and enhance generalization. Moreover, balancing the dataset resolved the poor performance on the negative class seen in unbalanced data, as evidenced by an increase in recall from 86% to 96%. This result further establishes the importance of the steps involved in preprocessing data, like merging datasets and over-sampling, toward robust training of the models.

The hybrid architecture, which combined BGRU and LSTM layers, was a critical factor in the model's success. While the BGRU layer captured contextual nuances in both directions, the LSTM layer retained long-term dependencies, enabling deeper understandings of textual inputs. This combination, as compared to single-layer architectures or traditional RNNs, yielded notable improvements, especially in handling complex datasets that were usually filled with sequential dependencies, like reviews laden with sarcasm or implicitly stating sentiments.

This fact is further supported by the comparison with alternative architectures, such as LSTM, CNN+LSTM, and GRU+LSTM. All models had very good performance metrics, but it was the hybrid architecture that outperformed others across balanced datasets. For example, the proposed model achieved an F1-score of 96% for both positive and negative classes, which showed that the performance of the model is well balanced, while alternative approaches did not demonstrate this quality. The robustness of the approach is underlined by the capacity to preserve the performance in the majority class, while significantly enhancing the minority class.

The HBGRU-LSTM model of opinion mining has immense scope for real-time applications in many areas:

- Social media monitoring: The model can be integrated with real-time social media analytics software to track and analyze customer sentiment, brand image, and public opinion trends in real time. This can be used by companies for proactive marketing and reputation management.
- E-commerce and customer review analysis: The model can enhance real-time product review analysis, allowing e-commerce sites to categorize and summarize customer opinions efficiently. This can assist businesses in product development, customer issue resolution, and recommendation optimization.

- Financial and stock market prediction: Based on real-time news articles, financial reports, and social media sentiment analysis, the model can produce forecasts of stock direction and economic trends to help investors and financial analysts make decisions.
- Patient and healthcare feedback systems: The model can be utilized to implement on patient review systems such that relevant feedback on hospitals, doctors, and treatment can be collected. This will assist healthcare professionals in enhancing the patient experience and medical treatment according to sentiment patterns.
- Customer service automation: The model can be integrated into chat-bots and virtual assistants to monitor customer sentiment in real-time, providing adequate responses and escalation of critical issues to human representatives whenever needed.
- Political and social sentiment analysis: Governments and policymakers can use the model to track public mood in real-time for policies, elections, and issues of national interest so that they can make evidence-based decisions.

## 7. Conclusions

The increasing volume of opinions shared on social media platforms only confirms the demand for effective opinion mining tools. Businesses can benefit a lot from such data in their quest to improve their products and services.

This paper underlined the effectiveness of a hybrid bidirectional GRU-LSTM architecture in sentiment analysis; this indeed constitutes a sea change from the traditional approaches. Contextual nuances and class imbalances are handled successfully for improved accuracy and generalization across diverse datasets.

The proposed model is enhanced in different performance measures like accuracy, recall, and F1-score. The model addresses class imbalance using techniques like oversampling for the aim of generating a class-balanced set of sentiment classes. From the experimental results, it can be seen that class imbalance needs to be addressed when performing sentiment analysis because the utilization of class-balanced data significantly enhanced the performance of the model, particularly in detecting negative sentiments

The model worked effectively on benchmark datasets such as IMDB and Amazon reviews, its versatility. Nonetheless, additional experimentation on domain datasets is recommended to unlock its full potential in specific domains such as healthcare and finance. The future can also witness a combination of transformer-based models, which have been useful in feature representation and accuracy improvement. Its application can also give insights into consumer attitudes that can be beneficial to businesses and public opinion. Consistent performance on benchmark datasets, such as IMDB and Amazon reviews, is a sign of its generalizability across domains. Nevertheless, the model must be tested with more domain-specific datasets to realize its full potential in special domains such as healthcare, education, and finance.

**Author Contributions:** Conceptualization, M.S.H., A.H. and M.A.; methodology, M.S.H., A.H.; software, M.S.H. and A.H.; validation, M.S.H., A.H. and M.A.; formal analysis, S.A.A. and E.A.; investigation, E.A.; resources, S.A.A. and M.A.; data curation, S.A.A., M.A. and E.A.; writing–original draft, S.A.A., M.A. and E.A.; writing–review and editing, A.H.; visualization, S.A.A. and E.A.; supervision, M.S.H.; project administration, M.S.H. All authors have read and agreed to the published version of the manuscript.

**Funding:** This research received no external funding

**Institutional Review Board Statement:** Not applicable.

**Informed Consent Statement:** Not applicable.

**Data Availability Statement:** Publicly available datasets were analyzed in this study. Data can be found here: https://www.kaggle.com/datasets/ebiswas/imdb-review-dataset, accessed on 1 October 2024 and https://www.kaggle.com/datasets/saurav9786/amazon-product-reviews, accessed on 1 October 2024.

**Conflicts of Interest:** The authors declare no conflicts of interest.

## Abbreviations

The following abbreviations are used in this manuscript:

| | |
|---|---|
| BCE | Binary Cross-Entropy |
| BGRU | Bidirectional Gated Recurrent Unit |
| BoW | Bag of Words |
| CNN | Convolutional Neural Networks |
| DL | Deep Learning |
| DT | Decision Trees |
| GRUs | Gated Recurrent Units |
| HBGRU | Hybrid Bidirectional Gated Recurrent Unit |
| LSTM | Long Short Term Memory |
| ML | Machine Learning |
| NB | Naïve Bayes |
| NLP | Natural Language Processing |
| ReLU | Rectified Linear Unit |
| RNN | Recurrent Neural Networks |
| SVM | Support Vector Machine |
| TF-IDF | Term Frequency-Inverse Document Frequency |